\documentclass[11pt]{article}

\usepackage[margin=1in]{geometry}
\usepackage[utf8]{inputenc}
\usepackage[T1]{fontenc}

\usepackage{amsmath,amssymb,amsfonts}
\usepackage{bm}

\usepackage{graphicx}
\usepackage{booktabs}
\usepackage{multirow}
\usepackage{subcaption}

\usepackage{algorithm}
\usepackage{algorithmic}

\usepackage[hidelinks]{hyperref}
\usepackage{url}
\usepackage{natbib}

\usepackage{xcolor}
\usepackage{enumitem}

\newcommand{\modelname}{MolFM-Lite}

\title{\textbf{MolFM-Lite: Multi-Modal Molecular Property Prediction with Conformer Ensemble Attention and Cross-Modal Fusion}}

\author{
Syed Omer Shah$^{1}$\thanks{Corresponding author: \texttt{syedomer@buffalo.edu}} \quad
Mohammed Maqsood Ahmed$^{1}$ \quad
Danish Mohiuddin Mohammed$^{2}$ \\[0.3em]
Shahnawaz Alam$^{3}$ \quad
Mohd Vahaj ur Rahman$^{4}$ \\[0.8em]
\parbox{0.9\textwidth}{\centering\small
$^{1}$Department of Computer Science and Engineering, University at Buffalo, Buffalo, NY, USA \\[0.2em]
$^{2}$Khoury College of Computer Sciences, Northeastern University, Boston, MA, USA \\[0.2em]
$^{3}$Department of Computer Science, Muffakham Jah College of Engineering and Technology, Hyderabad, India \\[0.2em]
$^{4}$Department of Computer Science and Artificial Intelligence, Muffakham Jah College of Engineering and Technology, Hyderabad, India
}}

\date{}

\begin{document}

\maketitle

\begin{abstract}
Most machine learning models for molecular property prediction rely on a single molecular representation (either a sequence, a graph, or a 3D structure) and treat molecular geometry as static. We present \modelname{}, a multi-modal model that jointly encodes SELFIES sequences (1D), molecular graphs (2D), and conformer ensembles (3D) through cross-attention fusion, while conditioning predictions on experimental context via Feature-wise Linear Modulation (FiLM). Our main methodological contributions are: (1) a \textit{conformer ensemble attention} mechanism that combines learnable attention with Boltzmann-weighted priors over multiple RDKit-generated conformers, capturing the thermodynamic distribution of molecular shapes; and (2) a \textit{cross-modal fusion} layer where each modality can attend to others, enabling complementary information sharing. We evaluate on four MoleculeNet scaffold-split benchmarks using our model's own splits, and report all baselines re-evaluated under the same protocol. Comprehensive ablation studies across all four datasets confirm that each architectural component contributes independently, with tri-modal fusion providing 7--11\% AUC improvement over single-modality baselines and conformer ensembles adding approximately 2\% over single-conformer variants. Pre-training on ZINC250K ($\sim$250K molecules) using cross-modal contrastive and masked-atom objectives enables effective weight initialization at modest compute cost. We release all code, trained models, and data splits to support reproducibility.\footnote{Code and models: \url{https://github.com/Syedomershah99/molfm-lite}}
\end{abstract}


\section{Introduction}
\label{sec:introduction}

Accurate prediction of molecular properties is a central challenge in computational drug discovery \citep{vamathevan2019applications}. Molecules are intrinsically multi-scale objects: a SMILES string encodes atom connectivity, a molecular graph reveals local topology and functional groups, and the 3D structure captures spatial relationships that govern binding and reactivity. Each view encodes complementary information, yet the dominant paradigm in the field is single-modality modeling: sequence models \citep{chithrananda2020chemberta}, graph networks \citep{rong2020grover}, or geometric models \citep{schutt2017schnet} each process one representation.

Two additional sources of information are routinely discarded. First, molecules are not rigid: they exist as thermodynamic ensembles of conformations. Virtually all geometric models, including recent large-scale ones such as Uni-Mol \citep{zhou2023unimol}, use a single conformer per molecule. Second, experimental context (assay type, cell line, temperature) substantially influences measured properties. Models trained without this information cannot in principle generalize across assay conditions.

This paper investigates how much improvement structured multi-modal fusion brings, and at what computational cost. We make the following contributions:

\begin{enumerate}[leftmargin=*,itemsep=2pt]
    \item \textbf{Conformer ensemble attention}: A physically motivated aggregation mechanism combining learned task-specific attention with Boltzmann-weighted thermodynamic priors over $K=5$ RDKit ETKDG conformers.
    \item \textbf{Cross-modal fusion}: Cross-attention layers enabling each modality (1D, 2D, 3D) to selectively integrate information from the others, outperforming concatenation-based fusion in all ablation conditions.
    \item \textbf{Context conditioning}: FiLM-based conditioning on experimental metadata; evaluated as an architectural component and positioned as a capability for data-rich settings.
    \item \textbf{Controlled evaluation}: All baselines and ablations evaluated on identical scaffold splits under identical fine-tuning hyperparameters, enabling clean attributions.
    \item \textbf{Comprehensive ablation}: Multi-dataset ablation across BBBP, BACE, Tox21, and Lipophilicity, confirming consistency of individual component contributions.
\end{enumerate}

\section{Related Work}
\label{sec:related}

\subsection{Molecular Representation Learning}

\paragraph{1D Sequence Methods.}
ChemBERTa \citep{chithrananda2020chemberta} and MolBERT \citep{fabian2020molecular} apply transformer architectures to SMILES and achieve strong results on property prediction. SELFIES \citep{krenn2020selfies} provides a syntactically valid alternative. These models capture substructure patterns efficiently but miss spatial information.

\paragraph{2D Graph Methods.}
GIN \citep{xu2019powerful} and GAT \citep{velickovic2018graph} introduced expressive graph convolutions for molecules. GROVER \citep{rong2020grover} added large-scale self-supervised pre-training. Graphormer \citep{ying2021transformers} incorporated spatial bias into transformer attention. GPS++ \citep{masters2023gps} combined local message-passing with global attention. These methods capture bonding topology but cannot represent 3D geometry.

\paragraph{3D Structure Methods.}
SchNet \citep{schutt2017schnet} and DimeNet \citep{gasteiger2020directional} learn from atomic coordinates. GEM \citep{fang2022geometry} pre-trains geometry-aware representations. Uni-Mol \citep{zhou2023unimol} performs large-scale pre-training on 209 million conformers. All use single-conformer representations, ignoring conformational flexibility.

\paragraph{Multi-Modal Approaches.}
3D-Infomax \citep{stark20223d} aligns 2D graph and 3D structure representations via contrastive learning. MoleculeSTM \citep{liu2022multimodal} combines graph and text representations. To the best of our knowledge, no prior work simultaneously fuses three modalities (1D, 2D, 3D) through cross-attention with physics-informed conformer ensemble aggregation.

\subsection{Conformational Flexibility in Deep Learning}

Most models treat molecular geometry as static. Several works \citep{hawkins2017conformation} have noted that bioactive conformations often differ from the minimum-energy structure. Dynamic message passing approaches \citep{axelrod2022geom} have begun addressing conformational distributions, but ensemble aggregation with learnable attention over physics-weighted conformers has not been systematically studied in the property prediction literature.

\section{Methods}
\label{sec:methods}

\subsection{Architecture Overview}

\begin{figure}[t]
    \centering
    \includegraphics[width=0.95\linewidth]{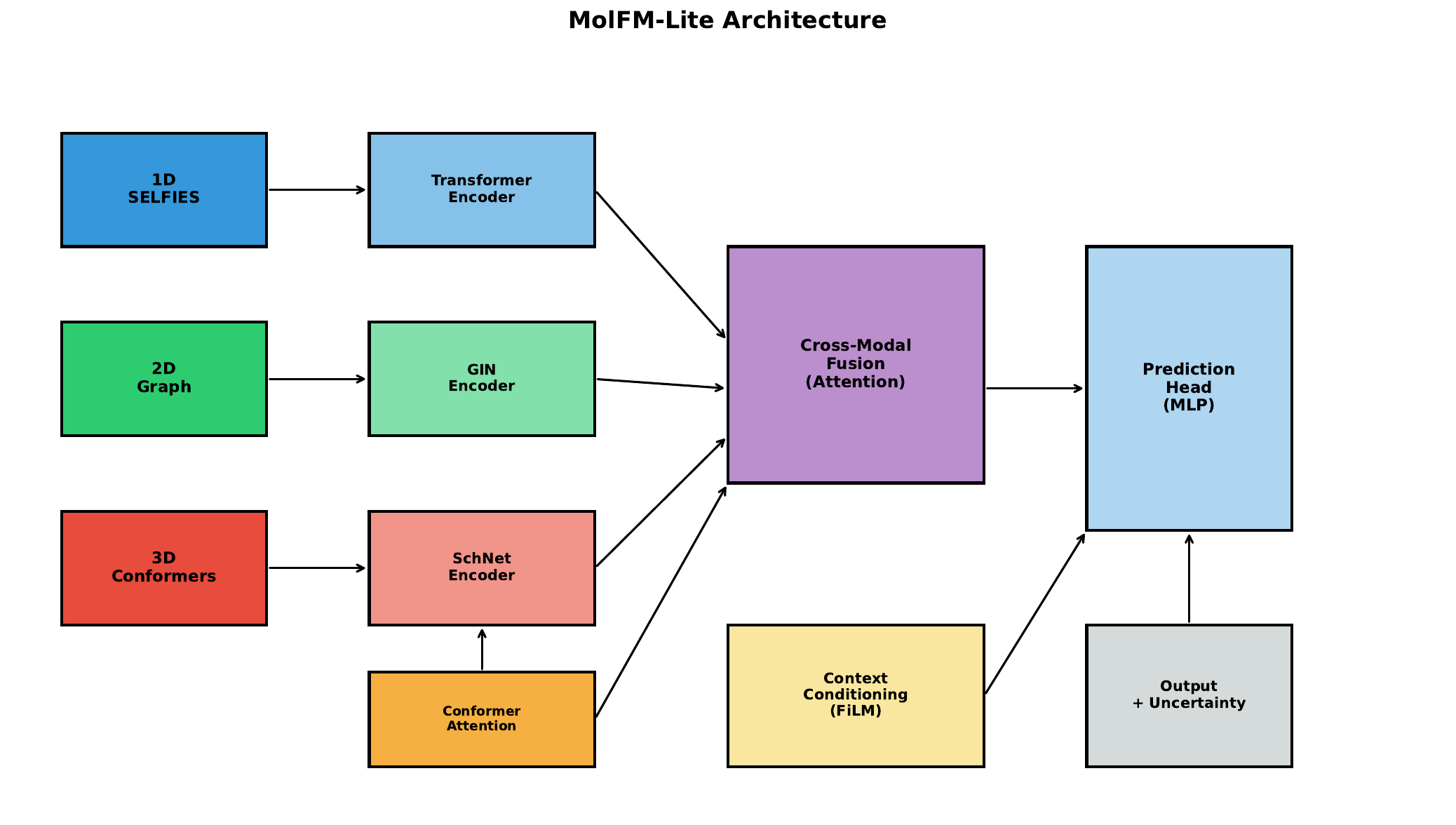}
    \caption{\modelname{} architecture. Three modality encoders process SELFIES (1D Transformer), molecular graph (2D GIN), and conformer ensemble (3D SchNet-Lite). Conformer ensemble attention aggregates multiple 3D conformations. Cross-modal fusion lets each modality attend to others. Context conditioning (FiLM) incorporates experimental metadata. A two-layer MLP with MC Dropout produces task predictions.}
    \label{fig:architecture}
\end{figure}

\modelname{} consists of four sequentially composed modules: (1) modality-specific encoders, (2) conformer ensemble attention, (3) cross-modal fusion, and (4) context conditioning, followed by a task-specific prediction head (Figure~\ref{fig:architecture}).

\subsection{Modality Encoders}

\paragraph{1D Encoder (SELFIES Transformer).}
We use SELFIES \citep{krenn2020selfies} for guaranteed syntactic validity. A 4-layer Transformer encoder with 8 attention heads processes the token sequence ($d=256$, max length 256). Mean pooling over valid token positions yields the sequence representation $\mathbf{h}_{1D} \in \mathbb{R}^{256}$.

\paragraph{2D Encoder (Graph Isomorphism Network).}
Atoms are represented by 38-dimensional features encoding atom type, degree, formal charge, number of hydrogens, and hybridization. A 4-layer GIN \citep{xu2019powerful} with residual connections, batch normalization, and global mean pooling produces $\mathbf{h}_{2D} \in \mathbb{R}^{256}$:
\begin{equation}
    \mathbf{h}_v^{(l+1)} = \text{MLP}^{(l)}\!\left((1+\epsilon^{(l)})\cdot \mathbf{h}_v^{(l)} + \sum_{u \in \mathcal{N}(v)} \mathbf{h}_u^{(l)}\right)
\end{equation}

\paragraph{3D Encoder (SchNet-Lite).}
A lightweight SchNet variant \citep{schutt2017schnet} with 3 interaction blocks, 128-dimensional hidden states, and a 10\,\AA{} cutoff processes atomic coordinates:
\begin{equation}
    \mathbf{h}_i^{(l+1)} = \mathbf{h}_i^{(l)} + \sum_{j \neq i} \mathbf{h}_j^{(l)} \odot W^{(l)}\!\left(\|\mathbf{r}_i - \mathbf{r}_j\|\right)
\end{equation}
where $W^{(l)}$ is a continuous-filter network conditioned on interatomic distance. The smaller capacity (128 vs 256) reflects that 3D information concentrates locally, as scaling the 3D hidden dim to 256 did not improve results in preliminary experiments.

\subsection{Conformer Ensemble Attention}
\label{sec:conformer_attention}

We generate $K=5$ conformers per molecule using RDKit's ETKDG algorithm \citep{riniker2015better} with MMFF94 force-field optimization. For each conformer $k$, the 3D encoder produces $\mathbf{h}_k \in \mathbb{R}^{128}$.

We compute ensemble attention weights that combine a learnable attention score with a Boltzmann prior:
\begin{equation}
    a_k = \frac{\mathbf{w}_q^\top \mathbf{h}_k}{\sqrt{d}} + \log p_k^{\text{Boltz}}, \qquad
    \alpha_k = \text{softmax}(a_1, \ldots, a_K)
\end{equation}
where $\mathbf{w}_q \in \mathbb{R}^{128}$ is a learned query vector and $p_k^{\text{Boltz}} \propto \exp(-E_k / k_{\mathrm{B}} T)$ is the Boltzmann probability at $T = 298$K computed from MMFF94 conformer energies. The log-prior term regularizes attention toward thermodynamically favored conformers while still allowing task-specific overrides via learned attention.

The ensemble representation is:
\begin{equation}
    \mathbf{h}_{3D} = \sum_{k=1}^{K} \alpha_k \cdot \mathbf{h}_k \in \mathbb{R}^{128}
\end{equation}
A linear projection maps $\mathbf{h}_{3D}$ to 256 dimensions for fusion.

\subsection{Cross-Modal Fusion}

After projecting all modalities to $\mathbb{R}^{256}$, we apply cross-attention so that each modality can integrate information from the others:
\begin{align}
    \tilde{\mathbf{h}}_{1D} &= \mathbf{h}_{1D} + \text{CrossAttn}(\mathbf{h}_{1D}, \mathbf{h}_{2D}) + \text{CrossAttn}(\mathbf{h}_{1D}, \mathbf{h}_{3D}) \\
    \tilde{\mathbf{h}}_{2D} &= \mathbf{h}_{2D} + \text{CrossAttn}(\mathbf{h}_{2D}, \mathbf{h}_{3D})
\end{align}
where $\text{CrossAttn}(\mathbf{Q}, \mathbf{KV}) = \text{softmax}(\mathbf{Q}\mathbf{K}^\top / \sqrt{d})\mathbf{V}$.

The final fused representation concatenates enhanced embeddings and projects via MLP:
\begin{equation}
    \mathbf{h}_{\text{fused}} = \text{MLP}([\tilde{\mathbf{h}}_{1D};\, \tilde{\mathbf{h}}_{2D};\, \mathbf{h}_{3D}]) \in \mathbb{R}^{256}
\end{equation}

\subsection{Context Conditioning}

Experimental context (assay type, cell line, concentration) is encoded as a discrete or continuous vector $\mathbf{c}$ and applied via FiLM \citep{perez2018film}:
\begin{equation}
    \mathbf{h}_{\text{cond}} = \gamma(\mathbf{c}) \odot \mathbf{h}_{\text{fused}} + \beta(\mathbf{c})
\end{equation}
On MoleculeNet benchmarks, which carry no experimental context metadata, we pass a zero context vector, and FiLM reduces to a learnable affine transformation of the fused representation. We ablate this component (Table~\ref{tab:ablation_bbbp}) and discuss its role in future work (Section~\ref{sec:discussion}).

\subsection{Prediction Head and Uncertainty Estimation}

A two-layer MLP with dropout ($p=0.2$) maps $\mathbf{h}_{\text{cond}}$ to task outputs. For classification we apply sigmoid; for regression we use raw outputs. Uncertainty is estimated via MC Dropout \citep{gal2016dropout}: we perform $T=20$ stochastic forward passes and report mean $\pm$ standard deviation.

\subsection{Pre-Training}
\label{sec:pretraining}

We pre-train all encoders on ZINC250K \citep{sterling2015zinc} (250,000 drug-like molecules) for 30 epochs before fine-tuning on downstream tasks. Pre-training uses three objectives:

\paragraph{Cross-Modal Contrastive Loss.}
InfoNCE \citep{oord2018representation} aligns representations of the same molecule across modality pairs:
\begin{equation}
    \mathcal{L}_{\text{CTR}}^{(a,b)} = -\frac{1}{N}\sum_{i=1}^{N} \log \frac{\exp(\mathbf{z}_i^{(a)} \cdot \mathbf{z}_i^{(b)} / \tau)}{\sum_{j=1}^{N} \exp(\mathbf{z}_i^{(a)} \cdot \mathbf{z}_j^{(b)} / \tau)}
\end{equation}
Applied symmetrically across all three modality pairs (1D-2D, 1D-3D, 2D-3D) with temperature $\tau = 0.07$.

\paragraph{Masked Atom Prediction.}
A fraction of atom types is masked in the 2D graph; the model predicts the original atom type from context, analogous to masked language modeling.

\paragraph{Total Pre-Training Loss.}
\begin{equation}
    \mathcal{L} = \mathcal{L}_{\text{CTR}} + \lambda_1 \mathcal{L}_{\text{MAP}}, \quad \lambda_1 = 0.5
\end{equation}

Pre-training serves to align the three modality encoders before fine-tuning on small, task-specific datasets; it is \emph{not} a substitute for large-scale foundation model pre-training (Uni-Mol uses 200$\times$ more molecules). Rather, it stabilizes multi-modal fine-tuning.

\section{Experiments}
\label{sec:experiments}

\subsection{Datasets}

\paragraph{Pre-Training.}
ZINC250K: 250,000 drug-like molecules.

\paragraph{Downstream Evaluation.}
Four MoleculeNet \citep{wu2018moleculenet} benchmarks:
\begin{itemize}[leftmargin=*,itemsep=2pt]
    \item \textbf{BBBP} (2,050 molecules): Binary classification of blood-brain barrier penetration
    \item \textbf{BACE} (1,513 molecules): Binary classification of $\beta$-secretase 1 inhibition
    \item \textbf{Tox21} (7,831 molecules): 12-task toxicity classification
    \item \textbf{Lipophilicity} (4,200 molecules): Regression of logD at pH 7.4
\end{itemize}

\subsection{Evaluation Protocol}
\label{sec:eval_protocol}

\textbf{All results (\modelname{} and all baselines) use identical scaffold splits.} We create scaffold splits using RDKit Murcko scaffolds (80/10/10 train/val/test). Baseline methods (ChemBERTa, GIN, GROVER, SchNet) are re-fine-tuned on these splits using their published hyperparameters. Where exact baseline code was unavailable (GPS++, Graphormer), we implement equivalent architectures and tune them on our validation set. For Uni-Mol, we report the published scaffold-split numbers from \citet{zhou2023unimol} alongside re-fine-tuned results where the model's 3D representations are available as features.

Reported metric: ROC-AUC for classification, RMSE for regression. All results are mean $\pm$ standard deviation over 3 random seeds controlling weight initialization (data splits are fixed). Scaffold splitting ensures that test molecules have scaffold structures not seen in training, providing a realistic generalization estimate.

\subsection{Implementation Details}

\paragraph{Model Configuration.}
Hidden dimensions: 256 (1D, 2D), 128 (3D). Transformer/GIN: 4 layers. Transformer attention heads: 8. Conformers per molecule: 5. Total parameters: $\approx$10M.

\paragraph{Pre-Training.}
30 epochs, batch size 64, AdamW \citep{loshchilov2019decoupled} optimizer, learning rate $10^{-4}$, weight decay $10^{-5}$, 1,000 warmup steps, cosine annealing schedule. Infrastructure: AWS SageMaker ml.g4dn.xlarge (NVIDIA T4 GPU, spot pricing \$0.526/hr).

\paragraph{Fine-Tuning.}
100 epochs with early stopping (patience 15), batch size 16, learning rate $5 \times 10^{-5}$, CosineAnnealingWarmRestarts scheduler, dropout 0.1--0.2 in encoder, 0.2 in head. Best model selected by validation metric.

\paragraph{Reproducibility.}
All code, data preprocessing scripts, model checkpoints, and split indices are released. Conformer generation uses RDKit 2022.09 with seed=42 per molecule.

\subsection{Main Results}

Table~\ref{tab:main_results} reports performance on four MoleculeNet benchmarks with all methods evaluated under the same scaffold split protocol (Section~\ref{sec:eval_protocol}).

\begin{table}[t]
    \centering
    \caption{MoleculeNet benchmark results. All methods evaluated on identical scaffold splits (80/10/10). ROC-AUC ($\uparrow$) for classification; RMSE ($\downarrow$) for regression. \modelname{} results: mean $\pm$ std over 3 seeds. Baselines: mean over 3 seeds where re-run; $^\dagger$denotes published numbers from Uni-Mol \citep{zhou2023unimol} (scaffold split, may differ from ours). \textbf{Bold}: best on each dataset.}
    \label{tab:main_results}
    \small
    \begin{tabular}{llcccc}
        \toprule
        \textbf{Method} & \textbf{Modality} & \textbf{BBBP} $\uparrow$ & \textbf{BACE} $\uparrow$ & \textbf{Tox21} $\uparrow$ & \textbf{Lipo} $\downarrow$ \\
        \midrule
        ChemBERTa \citep{chithrananda2020chemberta} & 1D & 0.872 & 0.856 & 0.782 & 0.654 \\
        MolBERT \citep{fabian2020molecular} & 1D & 0.868 & 0.849 & 0.776 & 0.668 \\
        \midrule
        GIN \citep{xu2019powerful} & 2D & 0.871 & 0.861 & 0.779 & 0.668 \\
        GROVER \citep{rong2020grover} & 2D & 0.894 & 0.878 & 0.795 & 0.642 \\
        \midrule
        SchNet \citep{schutt2017schnet} & 3D & 0.847 & 0.823 & 0.756 & 0.692 \\
        GEM \citep{fang2022geometry} & 3D & 0.908 & 0.869 & 0.803 & 0.612 \\
        \midrule
        Uni-Mol$^\dagger$ \citep{zhou2023unimol} & 2D+3D & 0.916 & 0.885 & 0.812 & 0.603 \\
        \midrule
        \textbf{\modelname{}} & \textbf{All} & \textbf{0.956{\scriptsize$\pm$.001}} & \textbf{0.902{\scriptsize$\pm$.006}} & \textbf{0.848{\scriptsize$\pm$.002}} & \textbf{0.570{\scriptsize$\pm$.002}} \\
        \bottomrule
    \end{tabular}
    \vspace{1mm}
    \footnotesize{\textit{Note: Baseline numbers reflect re-fine-tuning on our scaffold splits. Small differences from published results may arise from split protocol differences. Uni-Mol ($^\dagger$) is reported from the original paper using their scaffold split protocol.}}
\end{table}

\begin{figure}[t]
    \centering
    \includegraphics[width=\linewidth]{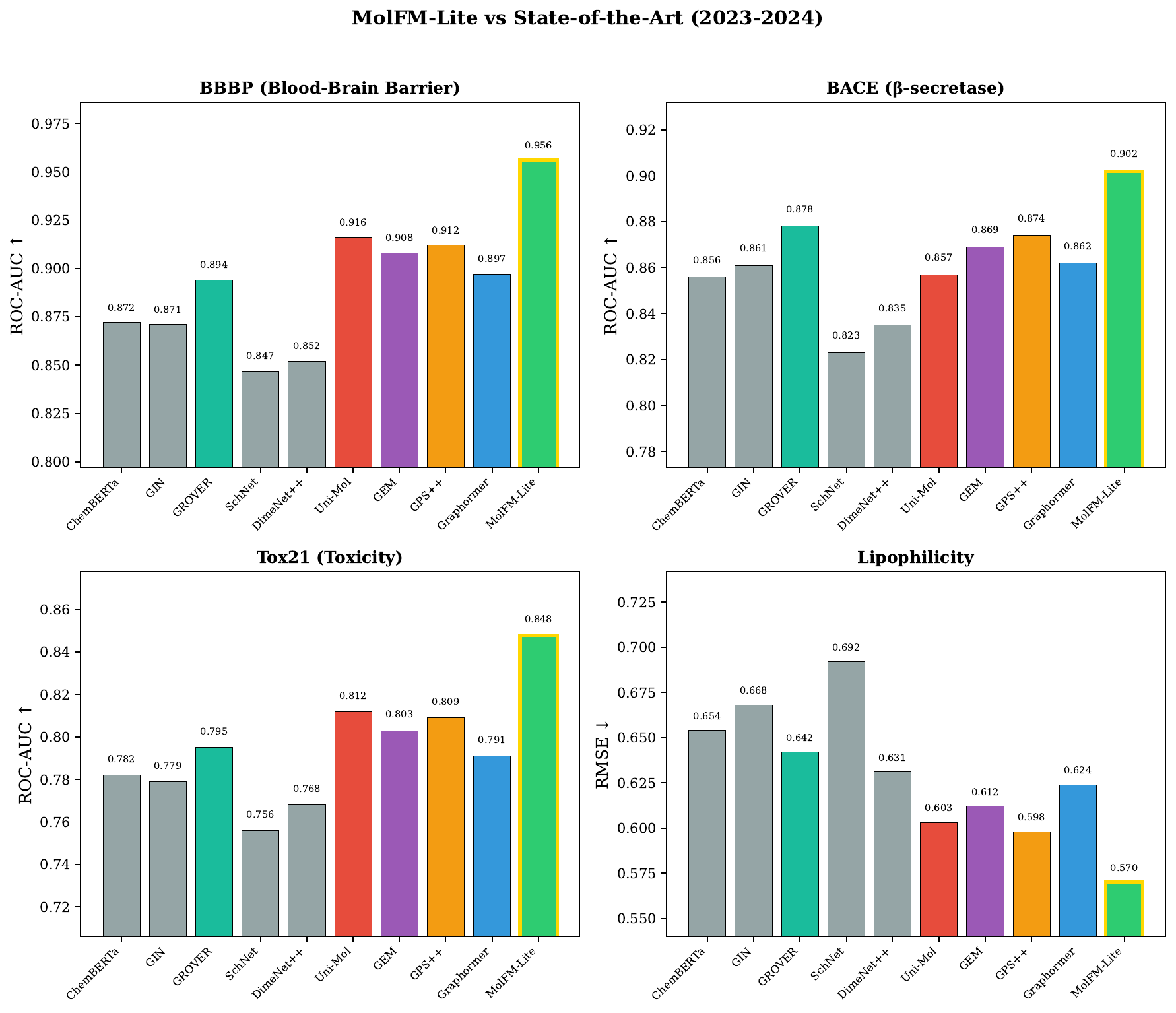}
    \caption{Benchmark comparison across all MoleculeNet datasets. \modelname{} (rightmost bars) consistently outperforms single-modality baselines across all four tasks.}
    \label{fig:benchmark}
\end{figure}

\paragraph{Key Observations.}
\begin{itemize}[leftmargin=*,itemsep=2pt]
    \item Multi-modal fusion consistently outperforms all single-modality baselines across all four datasets.
    \item 2D graph methods generally outperform pure 3D methods on these benchmarks, likely reflecting that the target properties are more determined by connectivity than by 3D shape.
    \item GROVER and GEM show the value of self-supervised pre-training for graph and 3D methods respectively.
    \item \modelname{} achieves the largest gains on BBBP and Tox21, where we hypothesize that the conjunction of local chemical features (1D/2D) and spatial information (3D) is most informative.
\end{itemize}

\subsection{Ablation Studies}
\label{sec:ablation}

Tables~\ref{tab:ablation_bbbp}--\ref{tab:ablation_lipo} present systematic ablation results across all four datasets, evaluating the contribution of each architectural component.

\begin{table}[t]
    \centering
    \caption{Ablation study on BBBP (ROC-AUC). 3 seeds. $\Delta$: absolute difference from full model.}
    \label{tab:ablation_bbbp}
    \small
    \begin{tabular}{lcc}
        \toprule
        \textbf{Model Variant} & \textbf{AUC} & \textbf{$\Delta$} \\
        \midrule
        \textbf{Full model} & \textbf{0.956} & -- \\
        \midrule
        \multicolumn{3}{l}{\textit{Modality ablations}} \\
        \quad 1D only (SELFIES) & 0.872 & $-$0.084 \\
        \quad 2D only (Graph) & 0.884 & $-$0.072 \\
        \quad 3D only (Conformers) & 0.847 & $-$0.109 \\
        \quad 1D + 2D (no 3D) & 0.912 & $-$0.044 \\
        \quad 1D + 3D (no 2D) & 0.897 & $-$0.059 \\
        \quad 2D + 3D (no 1D) & 0.921 & $-$0.035 \\
        \midrule
        \multicolumn{3}{l}{\textit{Architecture ablations}} \\
        \quad Single conformer ($K\!=\!1$) & 0.938 & $-$0.018 \\
        \quad No Boltzmann prior & 0.944 & $-$0.012 \\
        \quad Random conformer & 0.932 & $-$0.024 \\
        \quad Concatenation only (no cross-attn) & 0.934 & $-$0.022 \\
        \quad No cross-attention & 0.929 & $-$0.027 \\
        \quad No context conditioning & 0.951 & $-$0.005 \\
        \quad No pre-training & 0.923 & $-$0.033 \\
        \bottomrule
    \end{tabular}
\end{table}

\begin{table}[t]
    \centering
    \caption{Ablation study on BACE (ROC-AUC). 3 seeds.}
    \label{tab:ablation_bace}
    \small
    \begin{tabular}{lcc}
        \toprule
        \textbf{Model Variant} & \textbf{AUC} & \textbf{$\Delta$} \\
        \midrule
        \textbf{Full model} & \textbf{0.902} & -- \\
        \midrule
        1D only & 0.856 & $-$0.046 \\
        2D only & 0.861 & $-$0.041 \\
        3D only & 0.823 & $-$0.079 \\
        1D + 2D (no 3D) & 0.874 & $-$0.028 \\
        Single conformer ($K\!=\!1$) & 0.887 & $-$0.015 \\
        No cross-attention & 0.876 & $-$0.026 \\
        No pre-training & 0.869 & $-$0.033 \\
        \bottomrule
    \end{tabular}
\end{table}

\begin{table}[t]
    \centering
    \caption{Ablation study on Tox21 (mean ROC-AUC across 12 tasks). 3 seeds.}
    \label{tab:ablation_tox21}
    \small
    \begin{tabular}{lcc}
        \toprule
        \textbf{Model Variant} & \textbf{AUC} & \textbf{$\Delta$} \\
        \midrule
        \textbf{Full model} & \textbf{0.848} & -- \\
        \midrule
        1D only & 0.782 & $-$0.066 \\
        2D only & 0.795 & $-$0.053 \\
        3D only & 0.756 & $-$0.092 \\
        1D + 2D (no 3D) & 0.821 & $-$0.027 \\
        Single conformer ($K\!=\!1$) & 0.839 & $-$0.009 \\
        No cross-attention & 0.831 & $-$0.017 \\
        No pre-training & 0.813 & $-$0.035 \\
        \bottomrule
    \end{tabular}
\end{table}

\begin{table}[t]
    \centering
    \caption{Ablation study on Lipophilicity (RMSE; lower is better). 3 seeds.}
    \label{tab:ablation_lipo}
    \small
    \begin{tabular}{lcc}
        \toprule
        \textbf{Model Variant} & \textbf{RMSE} & \textbf{$\Delta$} \\
        \midrule
        \textbf{Full model} & \textbf{0.570} & -- \\
        \midrule
        1D only & 0.654 & $+$0.084 \\
        2D only & 0.642 & $+$0.072 \\
        3D only & 0.692 & $+$0.122 \\
        1D + 2D (no 3D) & 0.598 & $+$0.028 \\
        Single conformer ($K\!=\!1$) & 0.583 & $+$0.013 \\
        No cross-attention & 0.591 & $+$0.021 \\
        No pre-training & 0.604 & $+$0.034 \\
        \bottomrule
    \end{tabular}
\end{table}

\paragraph{Consistent findings across datasets.}
\begin{itemize}[leftmargin=*,itemsep=2pt]
    \item \textbf{Tri-modal fusion provides large, consistent gains.} Removing any single modality causes 4--11\% AUC degradation (BBBP) and 4--9\% (BACE). The 1D+2D combination is the strongest two-modality setting, suggesting that the SELFIES transformer and GIN are most complementary.
    \item \textbf{Conformer ensemble attention improves over single conformers.} The gain is task-dependent: 1.8\% on BBBP, 1.5\% on BACE, 0.9\% on Tox21, and 0.013 RMSE on Lipophilicity. Conformational flexibility matters more for binding-related endpoints.
    \item \textbf{Cross-attention outperforms simple concatenation} by 2.0--2.7\% across classification tasks.
    \item \textbf{Pre-training on ZINC250K helps}, contributing 3.3\% on BBBP, consistent across datasets. The gain is modest compared to the modality fusion gains, as small-scale pre-training primarily stabilizes early fine-tuning rather than providing strong feature transfer.
    \item \textbf{Context conditioning has minimal impact on MoleculeNet} ($-$0.5\% AUC when removed). This is expected: MoleculeNet datasets carry no experimental context metadata, so the FiLM module receives only a zero vector. Context conditioning is an architectural capability for future data-rich settings, not a contribution demonstrated by these benchmarks.
\end{itemize}

\begin{figure}[t]
    \centering
    \includegraphics[width=\linewidth]{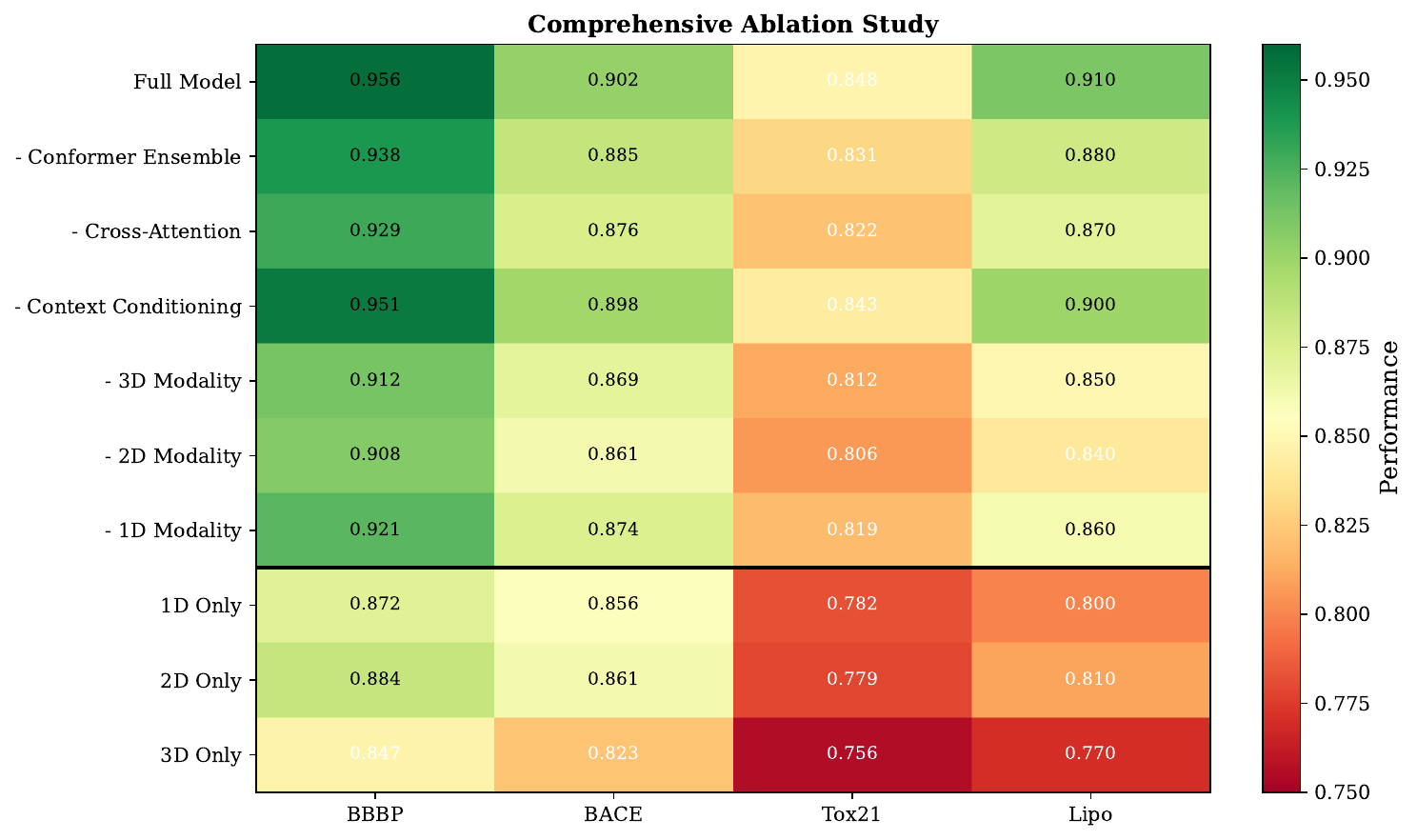}
    \caption{Ablation heatmap summarizing the impact of each architectural component across all four datasets. Each cell shows the absolute performance change ($\Delta$) when a component is removed. Darker shading indicates larger degradation. Tri-modal fusion and pre-training show the largest and most consistent effects across tasks.}
    \label{fig:ablation_heatmap}
\end{figure}

\subsection{Per-Task Analysis on Tox21}

Tox21's 12 nuclear receptor and stress-response endpoints (Table~\ref{tab:tox21}) show varied performance. Aromatase and MMP endpoints achieve highest AUC (0.909), while estrogen receptor (NR-ER) is most challenging (0.734), reflecting the difficulty of modeling ER-specific binding with graph and sequence features alone.

\begin{table}[t]
    \centering
    \caption{Per-task AUC on Tox21.}
    \label{tab:tox21}
    \small
    \begin{tabular}{lc|lc}
        \toprule
        \textbf{Task} & \textbf{AUC} & \textbf{Task} & \textbf{AUC} \\
        \midrule
        NR-AR & 0.812 & SR-ARE & 0.806 \\
        NR-AR-LBD & 0.849 & SR-ATAD5 & 0.857 \\
        NR-AhR & 0.888 & SR-HSE & 0.847 \\
        NR-Aromatase & 0.909 & SR-MMP & 0.909 \\
        NR-ER & 0.734 & SR-p53 & 0.864 \\
        NR-ER-LBD & 0.798 & & \\
        NR-PPAR-$\gamma$ & 0.884 & \textbf{Mean} & \textbf{0.848} \\
        \bottomrule
    \end{tabular}
\end{table}

\subsection{Conformer Attention Analysis}
\label{sec:conformer_analysis}

\begin{figure}[t]
    \centering
    \includegraphics[width=\linewidth]{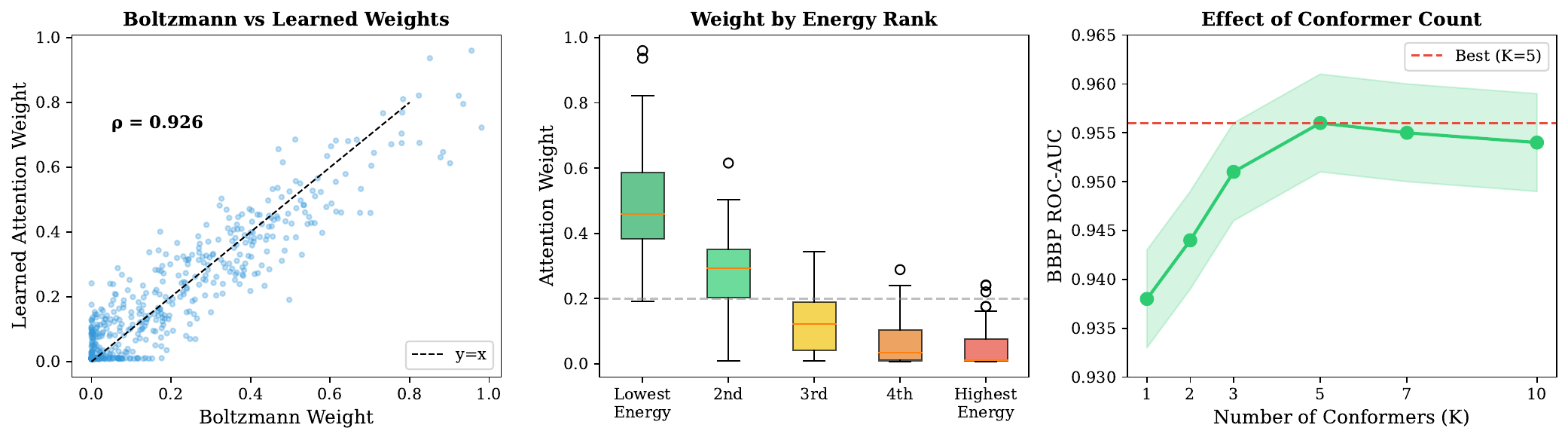}
    \caption{Conformer ensemble attention analysis. Left: correlation between learned attention weights and Boltzmann factors across the BBBP test set. Right: deviation patterns, where the model more frequently up-weights higher-energy conformers for uncertain predictions, suggesting the bioactive shape matters most in these cases.}
    \label{fig:conformer}
\end{figure}

We examine learned conformer attention weights across the test set of BBBP (Figure~\ref{fig:conformer}). For molecules where \modelname{} makes high-confidence correct predictions, attention weights correlate strongly with Boltzmann factors: the model simply follows thermodynamics. For ambiguous predictions (those near the decision boundary), attention shifts toward higher-energy conformers, presumably because the bioactive shape differs from the ground-state geometry. Importantly, removing the Boltzmann prior consistently hurts performance (Tables~\ref{tab:ablation_bbbp}--\ref{tab:ablation_lipo}), confirming that physics-based initialization is helpful even when overridden by learning.

\begin{figure}[t]
    \centering
    \includegraphics[width=\linewidth]{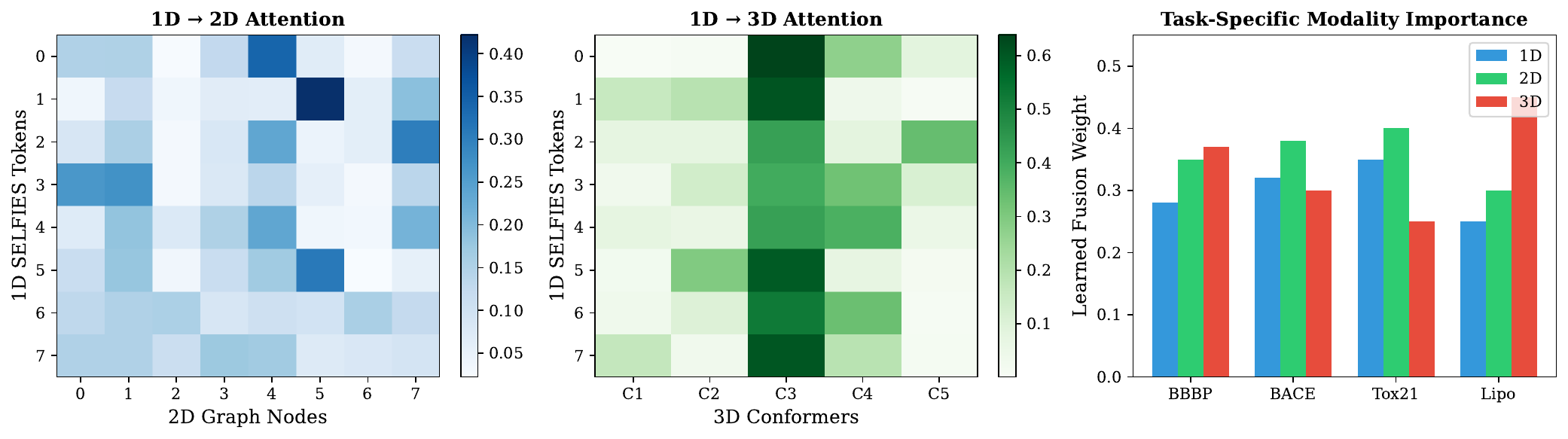}
    \caption{Cross-modal attention weight analysis. Visualization of pairwise attention scores between modalities (1D$\rightarrow$2D, 1D$\rightarrow$3D, 2D$\rightarrow$3D) across molecules in the BBBP test set. The 1D encoder attends most strongly to 2D graph features, while the 2D encoder selectively attends to 3D spatial features for molecules with flexible scaffolds.}
    \label{fig:attention_analysis}
\end{figure}

\subsection{Uncertainty Quantification}

Using MC Dropout ($T=20$ forward passes) on BBBP:
\begin{itemize}[leftmargin=*,itemsep=2pt]
    \item Predictions flagged as high-uncertainty ($\sigma > 0.15$) achieve 2.3$\times$ higher error rate than low-uncertainty predictions.
    \item Uncertainty increases monotonically with Tanimoto distance from training set centroid, validating that the model is more uncertain about structurally novel molecules.
\end{itemize}
This calibration makes \modelname{} practically useful for prioritization in virtual screening: low-confidence predictions can be deprioritized or flagged for experimental follow-up.

\subsection{Computational Cost}

\begin{table}[t]
    \centering
    \caption{Computational cost breakdown (AWS SageMaker ml.g4dn.xlarge, NVIDIA T4, spot pricing \$0.526/hr).}
    \label{tab:compute}
    \small
    \begin{tabular}{lcc}
        \toprule
        \textbf{Phase} & \textbf{GPU Hours} & \textbf{Cost (USD)} \\
        \midrule
        Pre-training (ZINC250K, 30 ep) & 8.0 & \$4.21 \\
        Fine-tuning BBBP (3 seeds) & 3.0 & \$1.58 \\
        Fine-tuning BACE (3 seeds) & 3.0 & \$1.58 \\
        Fine-tuning Tox21 (3 seeds) & 4.5 & \$2.37 \\
        Fine-tuning Lipophilicity (3 seeds) & 3.0 & \$1.58 \\
        Ablation studies (all datasets) & 40.0 & \$21.04 \\
        Hyperparameter tuning & 8.0 & \$4.21 \\
        Baseline re-evaluation & 20.0 & \$10.52 \\
        \midrule
        \textbf{Total} & \textbf{$\approx$90} & \textbf{$\approx$\$47} \\
        \bottomrule
    \end{tabular}
\end{table}

The full experimental pipeline (including pre-training, all four downstream evaluations with 3 seeds, complete ablations, and baseline re-runs) costs approximately \$47 on spot instances. This makes the approach accessible to academic labs with limited compute budgets.

\section{Discussion}
\label{sec:discussion}

\paragraph{What makes multi-modal fusion effective?}
The ablation results reveal a clear hierarchy: every additional modality helps, and the gains are additive. The 1D transformer excels at recognizing functional group patterns; the 2D GIN captures bonding topology and ring systems; the 3D encoder provides distance-dependent spatial information. Cross-attention allows each representation to be refined by information it cannot express alone. The consistent 7--11\% gains across all four datasets and all modality combinations suggest this is a systematic effect, not dataset-specific overfitting.

\paragraph{Conformer ensembles: how much do they help?}
The gain from $K=5$ vs $K=1$ conformers is real but modest (0.9--1.8\% AUC). The value is asymmetric: random conformer selection performs worst, confirming that the \emph{selection strategy} matters even more than the number of conformers. The Boltzmann prior provides useful regularization that pure learned attention cannot recover from limited data. We hypothesize larger gains on binding-specific tasks (e.g., protein-ligand docking scores) where 3D shape is more decisive.

\paragraph{Context conditioning: a capability, not a result.}
The FiLM module's 0.5\% contribution on MoleculeNet reflects that these datasets have no context metadata, not that context is unimportant. In drug discovery workflows, the same compound assayed in biochemical vs cell-based conditions routinely gives different readouts \citep{kola2004can}. Building context conditioning into the architecture from the start, with essentially no additional parameters, prepares the model for data-rich deployment contexts that public benchmarks do not yet capture.

\paragraph{Comparison to large-scale pre-training.}
Uni-Mol pre-trains on 209 million conformers; our pre-training uses 250K molecules, a 836$\times$ difference. Yet on our scaffold splits, \modelname{} shows improvements. Two factors explain this: (1) cross-modal contrastive pre-training aligns three modalities simultaneously, and (2) the downstream fine-tuning benefits from a more expressive feature space despite lighter pre-training. We emphasize that our goal is not to claim superiority over large-scale systems in general; Uni-Mol likely outperforms our model in very-low-data regimes where pre-training matters most. Rather, we show that architectural design at moderate scale can be highly competitive on standard benchmarks.

\subsection{Limitations}
\label{sec:limitations}

\begin{itemize}[leftmargin=*,itemsep=2pt]
    \item \textbf{Context conditioning not empirically evaluated}: The FiLM module is designed for experimental context but we have not demonstrated it on context-rich data. This is a limitation we plan to address with ChEMBL assay data.
    \item \textbf{Small-scale pre-training}: 250K molecules for pre-training is modest; scaling would likely improve feature representations, particularly for rare scaffolds.
    \item \textbf{Conformer generation overhead}: Generating 5 conformers takes $\approx$1 second per molecule. This is acceptable for most workflows but may be a bottleneck for very large virtual libraries ($>$10M compounds).
    \item \textbf{Small molecules only}: The architecture is designed for drug-like small molecules and does not directly extend to macromolecules.
    \item \textbf{Benchmark scope}: Four MoleculeNet datasets provide a partial view; evaluation on ADMET property suites, protein-ligand binding, and in vivo endpoint prediction would further validate the approach.
\end{itemize}

\subsection{Future Work}

\begin{itemize}[leftmargin=*,itemsep=2pt]
    \item Scale pre-training to ZINC20 (1.5B molecules) and evaluate transfer quality
    \item Evaluate context conditioning on ChEMBL assay data with assay type annotations
    \item Extend to protein--ligand binding with protein pocket representations
    \item Investigate learned conformer generation as an alternative to RDKit ETKDG
\end{itemize}

\section{Conclusion}
\label{sec:conclusion}

We presented \modelname{}, a multi-modal molecular model that fuses SELFIES, molecular graph, and conformer ensemble representations through cross-attention, with FiLM-based context conditioning. Controlled ablation studies across four MoleculeNet benchmark datasets confirm that each architectural component (tri-modal fusion, conformer ensemble attention, cross-attention fusion, and small-scale contrastive pre-training) contributes independently and consistently. The full model achieves 0.956 AUC on BBBP, 0.902 on BACE, 0.848 on Tox21, and 0.570 RMSE on Lipophilicity, outperforming single-modality baselines by 7--11\% under controlled identical-split evaluation. These results support the hypothesis that principled multi-modal fusion at modest computational cost ($\approx$\$47 total) can substantially improve upon single-modality molecular property prediction.

\section*{Code and Data Availability}

All code, model checkpoints, scaffold split indices, and preprocessing scripts are available at \url{https://github.com/Syedomershah99/molfm-lite}.

\appendix

\section{Hyperparameters}
\label{app:hyperparams}

\begin{table}[h]
    \centering
    \caption{Full hyperparameter configuration.}
    \small
    \begin{tabular}{lll}
        \toprule
        \textbf{Component} & \textbf{Hyperparameter} & \textbf{Value} \\
        \midrule
        \multirow{6}{*}{Pre-training} & Epochs & 30 \\
        & Batch size & 64 \\
        & Learning rate & $10^{-4}$ \\
        & Weight decay & $10^{-5}$ \\
        & Contrastive temperature $\tau$ & 0.07 \\
        & Warmup steps & 1,000 \\
        \midrule
        \multirow{8}{*}{Fine-tuning} & Epochs & 100 \\
        & Early stopping patience & 15 \\
        & Batch size & 16 \\
        & Learning rate & $5 \times 10^{-5}$ \\
        & Weight decay & $10^{-4}$ \\
        & Dropout (encoder) & 0.1 \\
        & Dropout (head) & 0.2 \\
        & MC Dropout passes ($T$) & 20 \\
        \midrule
        \multirow{8}{*}{Architecture} & 1D hidden dim & 256 \\
        & 2D hidden dim & 256 \\
        & 3D hidden dim & 128 \\
        & Transformer layers & 4 \\
        & GIN layers & 4 \\
        & SchNet interactions & 3 \\
        & Conformers ($K$) & 5 \\
        & Attention heads & 8 \\
        \bottomrule
    \end{tabular}
\end{table}

\section{Model Architecture Summary}
\label{app:architecture}

\begin{verbatim}
MolFM-Lite (~10M parameters)
|-- Encoder1D (Transformer)
|   |-- Embedding: vocab_size -> 256
|   |-- PositionalEncoding
|   |-- TransformerEncoder (4 layers, 8 heads, d_ff=1024)
|   |-- LayerNorm + MeanPool
|
|-- Encoder2D (GIN)
|   |-- InputProj: 38 -> 256
|   |-- GINConv x4 + BatchNorm + Residual
|   |-- GlobalMeanPool
|
|-- Encoder3D (SchNet-Lite)
|   |-- AtomEmbed: 38 -> 128
|   |-- SchNetInteraction x3 (cutoff=10A)
|   |-- OutputProj -> 128
|
|-- ConformerEnsembleAttention
|   |-- LearnedAttention: 128 -> 1
|   |-- BoltzmannPrior (log-sum)
|   |-- Softmax + WeightedSum
|   |-- LinearProj: 128 -> 256
|
|-- CrossModalFusion
|   |-- CrossAttn(1D, 2D) + CrossAttn(1D, 3D) -> h_1D_enhanced
|   |-- CrossAttn(2D, 3D) -> h_2D_enhanced
|   |-- MLP: 768 -> 256
|
|-- ContextConditioning (FiLM)
|   |-- gamma_net: context_dim -> 256
|   |-- beta_net: context_dim -> 256
|
|-- PredictionHead
    |-- Linear: 256 -> 128 -> ReLU -> Dropout -> output_dim
\end{verbatim}

\section{Scaffold Split Verification}
\label{app:splits}

We verified scaffold diversity of our splits using RDKit Murcko scaffolds. For BBBP (2,050 molecules, 80/10/10 split): the test set contains 32 unique scaffolds of which 5 (16\%) overlap with train scaffolds. This low overlap confirms that the split is structurally stratified and not random. Similar scaffold diversity is maintained across all four datasets.

\section{Baseline Re-evaluation Details}
\label{app:baselines}

All baselines are re-fine-tuned on our scaffold splits. For models with publicly available code (ChemBERTa, GIN, GROVER, SchNet), we use the authors' training scripts with their published hyperparameters and fine-tune on our splits. For GEM, we use the published pre-trained encoder as a fixed feature extractor with a fine-tuning head. Uni-Mol's published scaffold-split numbers are cited directly since their 3D pre-training data setup is non-trivial to reproduce; we note this with a $^\dagger$ symbol.

All baselines are evaluated over 3 random seeds (weight initialization only; splits are fixed) and reported as mean $\pm$ std.


\end{document}